\documentclass{article}

\usepackage{graphicx} 
\usepackage[utf8]{inputenc} 
\usepackage[T1]{fontenc}    
\usepackage{hyperref}       
\usepackage{url}            
\usepackage{booktabs}       
\usepackage{amsfonts}       
\usepackage{nicefrac}       
\usepackage{microtype}      
\usepackage{xcolor}         
\usepackage{algorithm}
\usepackage{algpseudocode}
\usepackage{subfigure}
\usepackage{arydshln}
\usepackage{makecell}
\usepackage{tabularray}
\usepackage[normalem]{ulem}
\usepackage{amsmath}
\usepackage{amssymb}
\usepackage{mathtools}
\usepackage{amsthm}
\usepackage{float}
\usepackage[many]{tcolorbox}
\tcbuselibrary{listings,skins}
\usepackage{multicol}
\usepackage{multirow}
\usepackage{longtable}
\usepackage{verbatim}
\usepackage{enumitem}
\usepackage{natbib}
\usepackage[final]{style}
\usepackage{listings}
\usepackage{xcolor}
\usepackage{soul}
\usepackage{ulem}
\usepackage{caption}  
\usepackage{hyperref}       
\usepackage{url}            
\usepackage{booktabs}       
\usepackage{amsfonts}       
\usepackage{nicefrac}       
\usepackage{microtype}      
\usepackage{xcolor}         
\usepackage{amsmath}

\usepackage{booktabs}
\usepackage{graphicx}
\usepackage{makecell}
\usepackage{wrapfig}
\usepackage{xcolor}

\usepackage{algorithm}
\usepackage{algpseudocode}
\usepackage{amssymb}
\usepackage{enumitem}
\usepackage{fvextra}

\definecolor{IceBlue}{HTML}{e0f7fa}
\definecolor{NavyBlue}{HTML}{000080}
\definecolor{darkbrown}{HTML}{8B4513}

\newtcolorbox{dialogbox}[1][]{
    colback=IceBlue,
    colframe=NavyBlue,
    fonttitle=\bfseries,
    title=#1,
    listing options={
        basicstyle=\small\ttfamily,
        breaklines=true,
        xleftmargin=0pt,
        xrightmargin=0pt,
        alsoletter={0123456789},
        morekeywords={Content1}, keywordstyle=\color{darkbrown}
    },
    verbatim
}

\usepackage{amsmath,amsfonts,bm, amssymb}
\usepackage{algorithm}      
\usepackage{algpseudocode}  

\def\eqref#1{equation~\ref{#1}}

\def\1{\bm{1}}

\DeclareMathAlphabet{\mathsfit}{\encodingdefault}{\sfdefault}{m}{sl}
\SetMathAlphabet{\mathsfit}{bold}{\encodingdefault}{\sfdefault}{bx}{n}

\theoremstyle{plain}

\theoremstyle{definition}

\theoremstyle{remark}

\title{GEAR: Granularity-Adaptive Advantage Reweighting \\ for LLM Agents via Self-Distillation}

\author{
Sijia Li$^{1,2}$\thanks{Corresponding authors.}  \quad
Yuchen Huang$^{1}$ \quad
Zifan Liu$^{1}$ \quad
Yanping Li$^{1}$ \quad
Jingjing Fu$^{2}$  \\  
\textbf{Li Zhao$^{2}$ \quad
Jiang Bian$^{2}$ \quad
Ling Zhang$^{2}$\thanks{Corresponding authors.} \quad
Jun Zhang$^{1}$\footnotemark[1]  \quad Rui Wang$^{2}$\footnotemark[1]} \\
$^{1}$Hong Kong University of Science and Technology \\
$^{2}$Microsoft Research Asia  \\
\texttt{\{slifg,yhuanggn,zliuft,ylitx\}@connect.ust.hk},
\texttt{eejzhang@ust.hk} \\
\texttt{\{ds.dashu, jiang.bian.prc, wrui0920\}@gmail.com}
\texttt{\{lizo, zhangling\}@microsoft.com}
}

\date{May 2025}

\definecolor{navyblue}{HTML}{1F77B4}
\definecolor{darkred}{HTML}{d62728}

\algrenewcommand\algorithmiccomment[1]{\hfill\textcolor{gray}{// #1}}

\begin{document}

\maketitle

\begin{abstract}

Reinforcement learning has become a widely used post-training approach for LLM agents, where training commonly relies on outcome-level rewards that provide only coarse supervision. While finer-grained credit assignment is promising for effective policy updates, obtaining reliable local credit and assigning it to the right parts of the long-horizon trajectory remains an open challenge.
In this paper, we propose \textbf{Granularity-adaptivE Advantage Reweighting} (GEAR), an adaptive-granularity credit assignment framework that reshapes the trajectory-level GRPO advantage using token- and segment-level signals derived from self-distillation. GEAR compares an on-policy student with a ground-truth-conditioned teacher to obtain a reference-guided divergence signal for identifying adaptive segment boundaries and modulating local advantage weights. This divergence often spikes at the onset of a semantic deviation, while later tokens in the same autoregressive continuation may return to low divergence. GEAR therefore treats such spikes as anchors for adaptive credit regions: where the student remains aligned with the teacher, token-level resolution is preserved; where it departs, GEAR groups the corresponding continuation into an adaptive segment and uses the divergence at the departure point to modulate the segment’s advantage.
Experiments across eight mathematical reasoning and agentic tool-use benchmarks with Qwen3 4B and 8B models show that GEAR consistently outperforms standard GRPO, self-distillation-only baselines, and token- or turn-level credit-assignment methods. The gains are especially strong on benchmarks with lower GRPO baseline accuracy, reaching up to around 20\% over GRPO, suggesting that the proposed adaptive reweighting scheme is especially useful in more challenging long-horizon settings.

\end{abstract}

\section{Introduction}
\label{sec:Introduction}

Large language models (LLMs) are increasingly deployed as agents for complex, multi-step tasks~\citep{schick2023toolformer,barua2024exploring,wang2023voyager}. These agents typically operate through multi-turn interactions with external environments, interleaving reasoning with tool use such as retrieval or code execution~\citep{yao2023reactsynergizingreasoningacting}. In such settings, correctness is often determined only at the end of an interaction through a verifiable outcome reward, making supervision naturally trajectory-level~\citep{qian2025toolrl,yao2024tau}. As a result, outcome-based reinforcement learning (RL) has become a standard approach for improving LLM agents~\citep{ouyang2022training}. In particular, group-based methods such as Group Relative Policy Optimization (GRPO)~\citep{shao2024deepseekmath} are commonly used to optimize whole trajectories by comparing multiple sampled trajectories under the same task.


However, with outcome-only rewards, group-based RL inherits a flat credit assignment problem: the final reward provides only a coarse trajectory-level signal, making it difficult to accurately attribute the final outcome to individual tokens or actions within the sequence~\citep{sutton1998reinforcement,lightman2023let}.
This issue is especially severe in agent settings, where trajectories unfold over many reasoning and tool-use steps, and early steps can shape the states and choices that follow much later, while multiple distinct paths can lead to the same final outcome. 
Applying a single trajectory-level reward across the sequence can therefore reinforce locally harmful actions in successful trajectories while penalizing useful intermediate behaviors in failed ones, producing noisy policy updates and slowing the acquisition of reliable long-horizon behaviors~\citep{zeng2025reinforcing, maempowering}.

This has motivated finer-grained credit assignment beyond trajectory-level rewards, but long-horizon agent trajectories make this difficult in two ways. First, an intermediate decision may affect the final outcome only through later interactions. A reasoning step, tool call, or observation integration can become useful only after subsequent observations and decisions, making it hard to access its contribution locally. Second, fine-grained credit assignment also depends on the unit over which credit is assigned. Turns and tool calls provide useful interfaces for organizing agent behavior, but they are not always the units over which policy updates should be shared. A decisive behavior may occur within a sub-turn phrase, a partial query, or a short reasoning transition, while neighboring tokens in the same turn play different roles. Thus, predefined units are useful approximations, but not necessarily a principled credit granularity.

Existing fine-grained methods still face limitations along both dimensions. Some methods derive local credit from intrinsic signals such as uncertainty, likelihood, or entropy~\citep{yu2025dapo, wang20258020rulehighentropyminority, wang2025information, fenggroup, peng2026hiper}, but these signals may capture policy-side distributional properties rather than actual task contribution. Stronger task-dependent signals, such as process rewards, rule-based feedback, or rollout-based estimates, can provide more direct supervision~\citep{qian2025toolrl,zeng2025reinforcing,maempowering,wang2024math,kazemnejad2024vineppo}, but they rely on intermediate evaluators or rollout estimates whose reliability depends on the task, evaluator quality, rollout coverage, and the current policy distribution. Meanwhile, many methods apply credit over predefined units, such as tokens, turns, tool calls, or heuristic segments. These units provide useful approximations to intermediate decision structure, but they may not match the behavioral unit that the local signal is meant to update. In tool-augmented agents, for example, a single turn may mix tool selection, argument construction, and observation interpretation, while the useful or harmful behavior may occupy only part of the turn. As a result, fine-grained credit assignment remains an open challenge for long-horizon agent RL.

We argue that effective local credit assignment requires coupling the credit signal with the behavioral unit over which it is applied. Motivated by this view, we propose \textbf{Granularity-adaptivE Advantage Reweighting} (GEAR), an adaptive-granularity credit assignment framework that uses an answer-conditioned self-distillation signal to derive local credit weights and identify credit boundaries that determine how these weights are shared. GEAR then uses the resulting weights to reshape the trajectory-level advantage, rather than introducing explicit step-level rewards. This allows GEAR to adapt both the strength and granularity of credit assignment without requiring additional step-level annotation or separately trained reward models.

Specifically, GEAR consists of three components. 
\textbf{(1)} \emph{Task-grounded adaptive credit assignment.}
GEAR constructs a task-grounded reference via self-distillation~\citep{zhao2026self}: conditioning the same model on the ground-truth-conditioned teacher policy, and the reverse Kullback–Leibler (KL) divergence between the student and teacher serves as a signal of behavioral deviation. A large reverse KL indicates that the student begins to depart from the task-correct continuation, naturally providing a useful signal for identifying the onset of meaningful credit regions. 
\textbf{(2)} \emph{Adaptive granularity.}
The empirical finding demonstrates that such divergence spikes often occur around transition markers (e.g., ``so'', ``but'', and ``therefore''), where the model commits to or shifts its reasoning path (Figure~\ref{fig:kl_tokens}). Since deviation typically extends beyond a single token into a coherent behavioral span, GEAR further uses token entropy as a boundary signal, as entropy spikes often indicate uncertainty or semantic transition, to determine segment extent. 
\textbf{(3)} \emph{Advantage redistribution.}
GEAR converts both token-level aligned regions and segment-level divergent regions into local credit weights that modulate the trajectory-level advantage. The reweighting is advantage-aware: in successful trajectories, high-divergence regions are down-weighted to avoid reinforcing unstable deviations, while in failed trajectories they receive stronger penalties. In this way, GEAR transforms coarse trajectory-level supervision into fine-grained credit assignment without requiring a separately trained process reward model.

We summarize the contributions of our work as follows:
\begin{itemize}[leftmargin=*, itemsep=2pt, topsep=2pt]
\item We propose \textbf{GEAR}, a granularity-adaptive credit assignment framework for outcome-based reinforcement learning in LLM agents. By using ground-truth-conditioned self-distillation to redistribute trajectory-level advantage, GEAR provides adaptive local supervision that remains anchored to final outcomes while avoiding explicit step-level rewards and separately trained reward models. This reduces reliance on task-specific intermediate supervision, making GEAR easy to adapt across outcome-based LLM RL settings.

\item We introduce an adaptive credit-region construction mechanism that makes trajectory segmentation signal-driven rather than predefined. Since reverse KL often spikes at the onset of a semantic departure, GEAR uses entropy to recover the corresponding continuation and share the high-deviation signal across the divergent region, while keeping low-divergence regions at token level. Compared with fixed-granularity methods, this avoids fragmented updates around meaningful deviations and artificial boundaries over the whole trajectory.

\item We validate GEAR through extensive experiments on eight mathematical reasoning and agentic tool-use benchmarks using Qwen3 4B and 8B models. GEAR consistently improves over GRPO and competing credit-assignment baselines, with particularly strong gains on benchmarks where the GRPO baseline accuracy is lower, reaching up to $\sim20$\% improvement. These results indicate that GEAR’s adaptive reweighting becomes especially valuable in challenging long-horizon settings.
\end{itemize}

\section{Related Work}
\label{sec:Related}

\subsection{Step-level Reward}

Post-training of LLMs relies on diverse reward signals with inherent trade-offs. Verifiable rewards (e.g., compilers and symbolic solvers) provide robust binary feedback for math and code \citep{guo2025deepseek, yu2025dapo}, but their coarse granularity exacerbates long-horizon credit assignment. In contrast, rewards from reward models offer nuanced, continuous signals but are prone to reward hacking, misspecification, and limited generalization \citep{li2024process, zhong2025comprehensive, xi2026agentprm}. LLM-as-a-judge approaches enable scalable, criteria-based evaluation for open-ended tasks \citep{wei2025reinforcing, zhang2026cm2}, but their feedback remains subjective and difficult to verify. To densify supervision, process-based and token-level methods introduce step-wise or lexical fine-grained signals \citep{guo2025segment, yang2026self, qian2025toolrl}. However, they often require expensive annotations or suffer from noise and context dependence, limiting their scalability and robustness.
In contrast, our method induces fine-grained adaptive credit assignment over dynamically formed segments and tokens by leveraging KL-based discrepancies between the same model's ground-truth-enhanced teacher distribution and its on-policy student distribution, thereby enabling scalable and structured advantage attribution without manual annotations or additional reward model training.


\subsection{Agentic RL}


Recent work on RL for LLM agents explores several aspects to improve long-horizon decision-making. Some methods redesign policy structures, such as decoupling planning and execution or separating high-level planning from low-level tool use~\citep{wang2025ragen, zhang2025agent}. Search-based methods incorporate lookahead mechanisms and LLM-guided value estimation to improve exploration and long-horizon optimization~\citep{yang2026tooltree, xietips, putta2024agent, zhou2023language}. Reward engineering densifies sparse outcome rewards into richer training signals~\citep{zhang2026cm2, feng2025retool}, while pure RL and self-play scale policy learning through large-scale environment interaction~\citep{zhai2025agentevolver, huang2025scaling}. Despite these advances, fine-grained credit assignment within trajectories remains largely underexplored.
In contrast, our method directly targets the RL credit assignment problem by introducing  an adaptive segment- and token-level mechanism that modulates trajectory-level advantages using the model's own predictive dynamics, without requiring explicit local reward engineering or additional architectural complexity.


\section{Preliminaries}
\label{sec:Preliminaries}

\subsection{Agent Execution}

We formulate agent execution as token-level sequential generation 
over a flattened interaction trace. Given a prompt $x$, the full trace is represented as a token sequence consisting of policy-generated tokens (reasoning, tool calls, and the final answer) and observation tokens returned by the environment (e.g., tool outputs). The state 
$s_t$ is the full prefix available before the next policy-generated 
token, including the prompt, previously generated tokens, and 
observation tokens. 
The action $a_t$ is the next policy-generated token sampled from $a_t \sim \pi_\theta(\cdot \mid s_t)$. Observation tokens are kept in the state but masked out from token-level training signals.

After termination, the trajectory receives a scalar outcome reward $R(\tau)$ measuring overall task success.
Since supervision is only available at the trajectory level, agentic RL naturally poses a long-horizon credit assignment problem. Importantly, all intermediate signals (e.g., reverse KL divergence and entropy) are defined at the token level, enabling fine-grained credit assignment over the trajectory.

\subsection{GRPO}
We consider policy optimization over trajectories generated by a policy $\pi_\theta$ conditioned on a prompt $x$. Following recent RL methods for LLM agents, we build upon the GRPO framework \cite{shao2024deepseekmath}. For each prompt $x$, we sample a group of $K$ trajectories:
$\{\tau^{(k)}\}_{k=1}^{K} \sim \pi_\theta(\cdot \mid x)$,
with corresponding rewards $\{R^{(k)}\}_{k=1}^K$. The advantage for each trajectory is computed by normalizing rewards within the group:
$A^{(k)} = \frac{R^{(k)} - \mu_R}{\sigma_R + \epsilon}$,
where $\mu_R$ and $\sigma_R$ denote the empirical mean and standard deviation of the group rewards. Following standard GRPO, the trajectory-level advantage is broadcast uniformly to all steps $A_t^{(k)} = A^{(k)}$ and optimized with the objective:
\begin{multline}
\label{eq:grpo_objective}
\mathcal{J}_{\text{GRPO}}(\theta) =
\mathbb{E}\Bigg[
\frac{1}{K}
\sum_{k=1}^{K}
\frac{1}{|\tau^{(k)}|}
\sum_{t=1}^{|\tau^{(k)}|}
\min\!\Big(
r_t^{(k)}(\theta) A_t^{(k)},
\mathrm{clip}\!\big(r_t^{(k)}(\theta),1-\epsilon,1+\epsilon\big) A_t^{(k)}
\Big)
\Bigg]  \\
- \beta D_{\mathrm{KL}}(\pi_\theta \| \pi_{\mathrm{ref}}),
\end{multline}
where $r_t^{(k)}(\theta)=\frac{\pi_\theta(a_t^{(k)}|s_t^{(k)})}{\pi_{\theta_{\mathrm{old}}}(a_t^{(k)}|s_t^{(k)})}$ is the policy ratio.
While GRPO yields a low-variance and stable advantage estimator, it assigns uniform credit across all actions within a trajectory, ignoring the heterogeneous contributions of individual actions and introducing noisy optimization in long-horizon trajectories. This motivates our adaptive, fine-grained credit refinement framework. 


\subsection{On-Policy Self-Distillation}
In on-policy RL, self-distillation~\citep{zhao2026self} encourages consistency between the policy's behavior under the state and under the ground-truth-conditioned state.
Given a trajectory $\tau=(s_t,a_t)_{t=1}^T \sim \pi_\theta$, a common reverse-KL objective is:
$\mathcal{L}_{\mathrm{distill}}(\theta)
=
\mathbb{E}_{\tau\sim\pi_\theta}
\left[
\sum_{t=1}^{T}
\left(
\log \pi_\theta(a_t\mid s_t)
-
\log \pi_\theta(a_t\mid s_t^\star)
\right)
\right]$,
where $s_t$ denotes the on-policy interaction state, and $s_t^\star$ denotes the ground truth-conditioned teacher state constructed from the same prompt under a reference trajectory $\tau^\star$ up to step $t$.

Rather than directly optimizing this objective, we repurpose the step-wise reverse KL signal as a fine-grained indicator of where the policy deviates from the ground-truth-conditioned behavior, and use it to shape the trajectory-level advantage for credit assignment.


\subsection{Policy Entropy}
To characterize the predictive uncertainty of the policy at each decision step, we consider the entropy of the action distribution induced by $\pi_\theta$. For a state $s_t$, the policy entropy is defined as
$H_t
=
\mathcal{H}\!\left(\pi_\theta(\cdot \mid s_t)\right)
=
-\sum_{a_t \in \mathcal{A}}
\pi_\theta(a_t \mid s_t)
\log \pi_\theta(a_t \mid s_t)$,
where $\mathcal{A}$ denotes the action space, including reasoning tokens, tool-use tokens, and final answer generation. Higher entropy indicates greater uncertainty, providing an auxiliary signal for differentiating action-level credit.

\section{Method}
\label{sec:method}

We propose \textbf{Granularity-AdaptivE Advantage Reweighting} (GEAR), a framework for fine-grained credit assignment in RL for fine-tuning LLM agents (Figure~\ref{fig:kl_tokens}). GEAR builds on GRPO-style optimization by introducing adaptive segment and token level reweighting of the trajectory-level advantage signal while preserving its low-variance property.



{Our method forms a unified pipeline for fine-grained credit assignment. 
It begins by estimating token-level behavioral deviation using reverse KL divergence between the teacher and student policies, producing advantage-aware token saliency signals (Section~\ref{ssec:Self_Credit_Assignment_Reverse_KL}). 
These signals are then jointly combined with token entropy to identify coherent behavioral segments, where reverse KL captures the onset of semantic deviation and entropy determines its extent, enabling adaptive segment-aware propagation of token credit within each segment while preserving original weights elsewhere (Section~\ref{ssec:entropy_seg}). 
Finally, the resulting fine-grained weights are used to modulate trajectory-level advantages according to their sign (Section~\ref{ssec:final_objective}), yielding adaptive credit assignment that is both locally precise and globally outcome-consistent.}


\subsection{Self-Distillation Credit Signal via Reverse KL Divergence}
\label{ssec:Self_Credit_Assignment_Reverse_KL}
Token-level credit assignment is challenging due to the lack of fine-grained supervision, as rewards are typically only available at the trajectory level. To address this, we leverage on-policy self-distillation~\citep{zhao2026self} to derive dense token-level signals. Specifically, we compute a discrepancy signal between teacher and student policies, which is then used to define an advantage-aware token-level weight.
The discrepancy signal is defined as the token-wise reverse KL divergence between the student policy and 
the ground-truth-conditioned teacher reference policy: 
\begin{equation}
\mathrm{rKL}_t
=
\log \pi_\theta(a_t\mid s_t)
-
\log \pi_\theta(a_t\mid s_t^\star),
\end{equation}
where $s_t^\star$ denotes the ground truth-conditioned state.
For stability and comparability, we normalize $\mathrm{rKL}_t$ to $[0,1]$ within each trajectory as $\widetilde{\mathrm{rKL}}$. A larger $\widetilde{\mathrm{rKL}}_t$ indicates greater deviation from the teacher distribution, implying that the corresponding token is farther from the ground-truth-guided signal and is less likely to contribute positively to the final outcome.

\subsection{Divergence-Aware Adaptive Segmentation}
\label{ssec:entropy_seg}

\begin{figure}[t]
    \centering
    \includegraphics[width=0.98\linewidth]{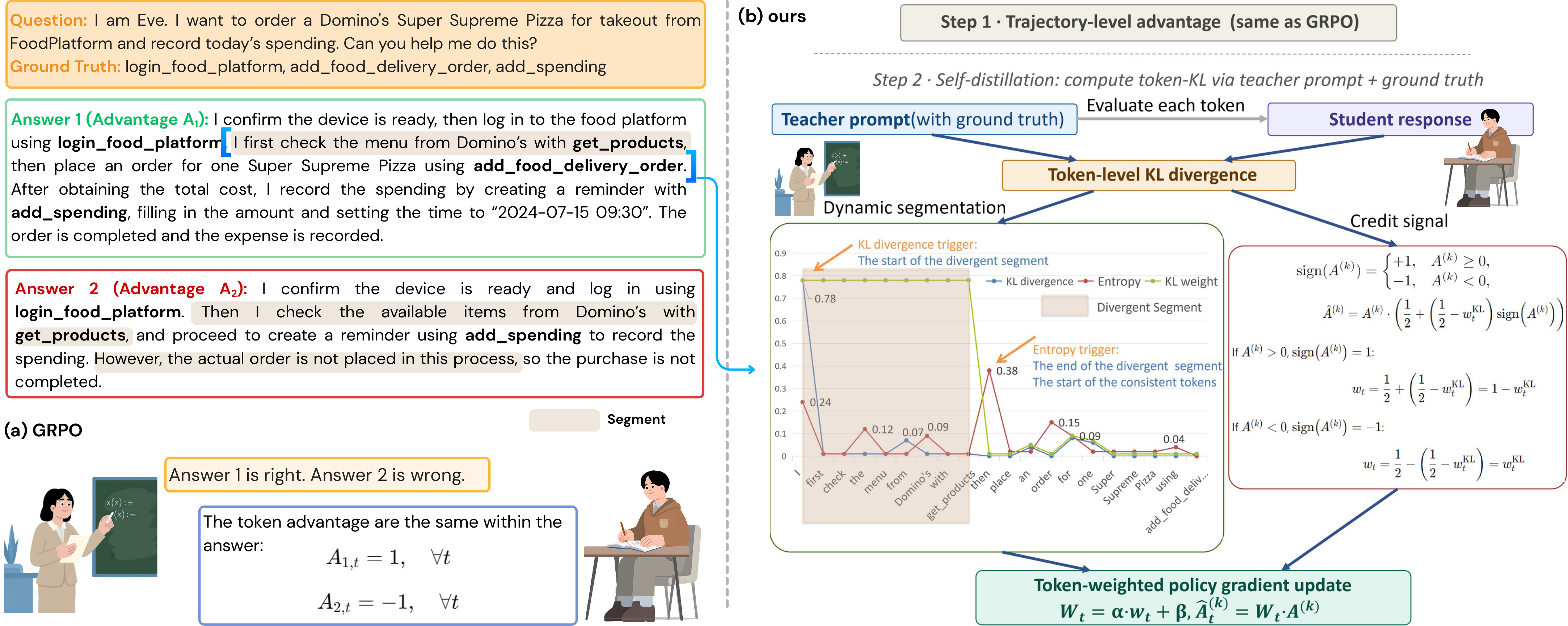}
    \caption{
    Illustration of \textbf{GEAR} for fine-grained credit assignment in agent RL. 
(a) GRPO assigns the same trajectory-level advantage to all tokens.
(b) GEAR preserves this trajectory-level advantage while redistributing credit at a finer granularity.
It computes token-wise reverse KL divergence between the student and a ground-truth–conditioned teacher, 
then uses KL peaks to identify segment onsets
and entropy to determine segment extent.
Within each divergent segment, all tokens share the onset credit, propagating a localized discrepancy signal over coherent spans. For consistent regions with low KL, token-level reverse KL is used directly. The resulting signals are then modulated by a sign-aware weighting function to produce the final token-level credit.}
    \label{fig:method}
    \vspace{-9pt}
\end{figure}

Directly applying discrepancy-based token-level weights to the trajectory-level advantage can induce high variance, as divergence signals are often sparse and concentrated on a few isolated tokens, particularly near the onset of a new reasoning step, resulting in noisy and unstable gradient updates.

To better understand this phenomenon,
we analyze 100 trajectories sampled from math tasks and report the 20 most frequent tokens whose normalized reverse-KL exceeds 0.1 ($\widetilde{\mathrm{rKL}}_t>0.1$; Appendix Figure~\ref{fig:kl_tokens}).The threshold is chosen based on empirical observations, as values above this level typically correspond to noticeable policy deviations.
High-KL tokens are predominantly discourse markers and transition phrases (e.g., ``So'', ``Let'', ``But'' and ``Therefore''), which typically indicate reasoning shifts, hypothesis revision, or structural decomposition. This suggests that elevated KL values are not uniformly distributed, but instead cluster around critical reasoning transitions.

Motivated by this observation, we introduce \emph{divergence-aware adaptive segmentation}, which groups localized high-divergence regions into coherent segments and enables structured segment-level advantage modulation, thereby reducing variance while preserving informative token-level discrepancy signals.

Specifically, we segment each trajectory using two complementary signals. The normalized divergence signal $\widetilde{\mathrm{rKL}}_t$ identifies token steps where the student policy substantially deviates from the teacher, while token-level entropy determines how far such deviations extend into subsequent steps.

Concretely, a new segment is initiated at token step $t_{\text{start}}$ whenever the normalized divergence exceeds a predefined threshold:
$\widetilde{\mathrm{rKL}}_{t_{\text{start}}}
>
\lambda_{\mathrm{KL}}.$
Once triggered, the segment continues until the first token step $t_{\text{end}}$ whose entropy exceeds a multiple of the entropy at the segment onset:
$H_{t_{\text{end}}}
>
\lambda_H H_{t_{\text{start}}}.$
All steps from $t_{\text{start}}$ to $t_{\text{end}}$ form a segment. The process then resumes from token step $t_{\text{end}}+1$, yielding a set of non-overlapping segments $\{S_n\}_{n=1}^{N}$ over the trajectory.

Based on this partition, we perform segment-level piecewise-constant reweighting. For all tokens within a segment $S_n$, we assign the same weight as the onset token at position $t_n$:
\begin{equation}
w_t^{\mathrm{KL}} =
\begin{cases}
\widetilde{\mathrm{rKL}}_{t_n}, & t \in S_n,\\
\widetilde{\mathrm{rKL}}_t, & \text{otherwise}.
\end{cases}
\end{equation}
{In this way, we partition the token sequence within each trajectory into divergent segments and consistent regions.  For divergent segments, all tokens share the credit signal from the segment onset, propagating localized discrepancy over a coherent span.  For consistent regions with persistently low KL divergence, tokens retain their own token-level weights to preserve fine-grained credit signals. This piecewise design smooths isolated KL spikes and yields more stable credit assignment.}

\subsection{Final Objective}
\label{ssec:final_objective}


Building upon the divergence-aware step-wise segment weight, we introduce an advantage-sign-aware token-level weighting scheme that adapts according to the sign of the advantage:
\begin{equation}
w_t
=
0.5+\bigl(0.5-w_t^{\mathrm{KL}}\bigr)\cdot \mathrm{sign}(A^{(k)}).
\end{equation}
This formulation induces an asymmetric and outcome-conditioned 
modulation. When the trajectory is globally beneficial ($A^{(k)} > 0$), 
tokens with large divergence from the teacher are \emph{downweighted}, 
suppressing potentially unreliable updates that may not have contributed 
to the positive outcome. Conversely, when the trajectory is harmful 
($A^{(k)} < 0$), such tokens are \emph{upweighted}, amplifying corrective 
gradients and discouraging the policy from repeating behaviorally 
deviant actions.

The final token-level weight is obtained via affine rescaling:
\begin{equation}
    W_t = \alpha \cdot w_t + \beta,
\end{equation}
where $\alpha > 0$ and $\beta$ are scalar hyperparameters controlling the dynamic range and offset of the modulation, respectively. We set $\beta = 1 - 0.5·\alpha$, such that the average modulation weight remains close to $1$, preserving the overall scale of the trajectory-level advantage while allowing adaptive up- and down-weighting across tokens. We then replace the uniformly assigned trajectory-level advantage $A_t^{(k)}$ in the GRPO objective (Eq. \ref{eq:grpo_objective}) with the reweighted token-level advantage $\hat{A}^{(k)}_t = W_t A^{(k)}$.
By retaining the trajectory-level advantage as the base signal and 
introducing only multiplicative modulation, our method preserves the 
low-variance characteristics of GRPO-style optimization while 
achieving fine-grained, outcome-aware credit assignment across 
reasoning steps---without requiring step-level reward annotations 
or auxiliary value networks.

\section{Experiments}
\label{sec:experiments}

\subsection{Experimental Settings}
\label{ssec:Experimental_Settings}



We train and evaluate our method along two complementary dimensions: \emph{agentic tool-use benchmarks}, which evaluate multi-step function-calling fidelity, and \emph{tool-integrated mathematical reasoning benchmarks}, which assess multi-step problem solving with access to a Python tool. For training, we use 5K mathematical reasoning examples from ARPO~\citep{dong2025agentic} and 4K agentic tool-use examples from ToolRL~\citep{qian2025toolrl}, following their original train-test splits.

\textbf{Mathematical Reasoning Benchmarks.}
Following prior work on tool-integrated reasoning~\cite{dong2025agentic}, we evaluate on standard mathematical reasoning benchmarks, including MATH~\cite{hendrycks2021measuring}, GSM8K~\cite{cobbe2021training}, and competition-level subsets such as MATH500~\cite{hendrycks2021measuring} and AIME 2024/2025. We report accuracy for each benchmark and adopt \textsc{Mean@16} across all tasks as the primary overall performance metric, setting the temperature and top-p to 0.6 and 0.95, respectively.

\paragraph{Agentic Tool-Use Benchmarks.}
We evaluate multi-step tool use on three benchmarks. \textbf{ToolSandbox}~\cite{lu2025toolsandbox} is a stateful benchmark with mutable world state; we use the Multiple\_Tool\_Call subset and evaluate milestone completion. \textbf{BFCL}~\cite{patil2025berkeley} evaluates multi-function calling (parallel, multiple, and parallel-multiple) using Abstract Syntax Tree (AST) matching. \textbf{ACEBench}~\cite{chen2025acebench} assesses agentic multi-step execution, where we report \textit{End-to-End Success} (exact match of all target attributes) and \textit{Process Accuracy} ($n/m$), measuring alignment with the reference function-call sequence.


\paragraph{Baselines.}
We compare our method against the following baselines. \textbf{OPSD}~\cite{zhao2026self} is an on-policy self-distillation framework in which a single model serves as both teacher and student under different contexts. \textbf{OPSD+RL} further incorporates OPSD as an auxiliary loss into the RL objective. \textbf{GRPO}~\cite{shao2024deepseekmath} is a representative outcome-only RL method that estimates advantages by normalizing terminal rewards within rollout groups. \textbf{ARPO}~\cite{dong2025agentic} improves multi-turn exploration by adaptively branching on token entropy spikes after tool-call feedback. \textbf{MT-GRPO}~\cite{zeng2025reinforcing} extends GRPO to multi-turn settings via turn-level reward for improved long-horizon credit assignment.

\paragraph{Implementation Details.}


We instantiate GEAR in a GRPO framework with \textsc{Qwen3-4B-Base} and \textsc{Qwen3-8B-Base}. We set $\lambda_{\mathrm{KL}}=0.1$, $\lambda_H=1.5$, $\alpha=0.2$ and $\beta=0.9$. For general-domain assessment, we additionally train a separate model on mixed agentic tool-use and mathematical reasoning data under the same training settings.

\subsection{Main Results}
\label{sec:main_results}
\paragraph{Agent Tool-Use.}
Table~\ref{tab:main_results} shows that our method consistently outperforms strong baselines on both Qwen3-4B and Qwen3-8B across agentic tool-use benchmarks. Compared with outcome-level RL methods such as GRPO, as well as recent multi-turn extensions including ARPO and MT-GRPO, our approach achieves the best overall performance across nearly all metrics, with particularly large gains on challenging benchmarks such as ToolSandbox. We attribute these gains to GEAR’s more effective credit assignment mechanism. ARPO relies on entropy-based branching signals, which are often noisy and fluctuate locally, leading to unstable and fragmented signals (Appendix Figure~\ref{fig:experiment_vis}). MT-GRPO introduces step-level rewards for finer supervision, but these manually designed signals are often task-specific and may hurt generalization out of domains.

By grounding local credit assignment in reverse-KL-based behavioral deviation while adaptively combining entropy only as a boundary signal, our method provides more stable and task-aligned fine-grained supervision without requiring explicit process rewards. As a result, these improvements remain consistent across different model scales and also transfer effectively to cross-domain data, where our method still yields gains over GRPO. This demonstrates that our fine-grained credit assignment mechanism is not specialized to a particular task domain or training setting, but instead exhibits strong generalization ability across model capacities, data distributions, and task domains.

\paragraph{Mathematical Reasoning.}
Table~\ref{tab:main_results} compares all baselines on mathematical reasoning benchmarks. We find that AIME24 and AIME25 better reflect model generalization ability than benchmarks such as GSM8K and MATH, as they involve more complex compositional reasoning and stronger distribution shifts, making them more sensitive to cross-domain transfer rather than in-domain memorization or pattern fitting.

Our method achieves a strong balance between in-domain performance and generalization. While maintaining competitive results on standard in-domain benchmarks (e.g., GSM8K and MATH), it consistently improves performance on AIME24 and AIME25, indicating that the proposed fine-grained credit assignment enhances reasoning robustness under distribution shift without sacrificing in-domain performance. In contrast, GRPO improves in-domain performance but shows noticeable degradation on AIME24 and AIME25, suggesting that optimizing solely for outcome-level rewards may lead to overfitting to in-domain patterns and weaker generalization under distribution shift.

Notably, the cross-domain trained variant of our model further improves AIME24 and AIME25 performance, even surpassing models trained directly on mathematical data. This indicates that our approach generalizes well across training domains and learns more transferable reasoning representations. Overall, these results show that our method improves both generalization and in-domain performance, demonstrating that GEAR is not task-specific but instead learns transferable reasoning capabilities that generalize to unseen tasks.

\begin{table*}[t]
\centering
\small
\caption{
Performance (\%) on agentic tool-use and mathematical reasoning benchmarks.
\textbf{Agent}: TS-M = ToolSandbox Multi-Tool,
BFCL Avg. = average over BFCL Parallel-Multiple / Parallel / Multiple,
ACE-E = ACEBench End-to-End accuracy,
ACE-P = ACEBench Process accuracy.
\textbf{Math}: Test(in domain), A24 = AIME24, A25 = AIME25, M500 = MATH500, GSM = GSM8K, MH = MATH.
Best results are \textbf{bolded}. 
}
\label{tab:main_results}

\vspace{1.5em}
\renewcommand{\arraystretch}{0.98}
\setlength{\tabcolsep}{3pt}

\resizebox{0.95\textwidth}{!}{%
\begin{tabular*}{\textwidth}{@{\extracolsep{\fill}}lcccc|cccccc@{}}
\toprule

& \multicolumn{4}{c|}{\textbf{Agentic Tool-Use}}
& \multicolumn{6}{c}{\textbf{Mathematical Reasoning}} \\

\cmidrule(lr){2-5}
\cmidrule(lr){6-11}

\textbf{Method}
& \textbf{TS-M}
& \textbf{BFCL Avg.}
& \textbf{ACE-E}
& \textbf{ACE-P}
& \textbf{Test}
& \textbf{A24}
& \textbf{A25}
& \textbf{GSM}
& \textbf{MH} 
& \textbf{M500}
\\

\midrule
\multicolumn{11}{l}{\textit{Qwen3-4B}}\\
Base      & 0.304 & 0.907 & 0.265 & 0.302 & 0.565 & 0.458 & 0.512 & 0.881 & 0.774 & 0.725 \\

OPSD      & 0.279 & 0.720 & 0.225 & 0.265 & 0.698 & 0.363 & 0.421 & 0.937 & 0.861 & 0.783 \\
OPSD+RL   & 0.337 & 0.871 & 0.243 & 0.281 & 0.668 & 0.396 & 0.458 & 0.949 & 0.848 & 0.746 \\
GRPO      & 0.346 & 0.919 & 0.312 & 0.356 &  0.676 & 0.431 & 0.474 & 0.924 & 0.836 & 0.754 \\
ARPO      & 0.367 & 0.923 & 0.236 & 0.262 & 0.691 & 0.463 & 0.489 & 0.932 & 0.852 & 0.764  \\
MT-GRPO   & 0.332 & 0.924 & 0.328 & 0.359 & 0.680 & 0.423 & 0.456 & 0.931 & 0.844 & 0.762 \\

\textbf{GEAR (Ours)}
& \textbf{0.423}
& \textbf{0.941}
& \textbf{0.363}
& \textbf{0.389}
& \textbf{0.738} & \textbf{0.503} & \textbf{0.534} & \textbf{0.958} & \textbf{0.879} & \textbf{0.802} \\

\midrule
\multicolumn{11}{l}{\textit{Qwen3-8B}}\\
Base      & 0.358 & 0.921 & 0.315 & 0.357 & 0.572 & 0.471 & 0.523 & 0.896 & 0.786 & 0.721 \\
OPSD      & 0.312 & 0.764 & 0.265 & 0.288 & 0.741 & 0.381 & 0.347 & 0.931 & 0.816 & 0.728 \\

OPSD+RL   & 0.368 & 0.885 & 0.305 & 0.329 & 0.701 & 0.471 & 0.445 & 0.925 & 0.804 & 0.719 \\
GRPO      & 0.383 & 0.920 & 0.313 & 0.358 & 0.755 & 0.465 & 0.463 & 0.932 & 0.812 & 0.728 \\
ARPO      & 0.379 & 0.926 & 0.325 & 0.363 & 0.763 & 0.479 & 0.478 & 0.928 & 0.822 & \textbf{0.742} \\
MT-GRPO   & 0.365 & 0.923 & 0.337 & 0.378 & 0.768 & 0.455 & 0.446 & 0.927 & 0.818 & 0.724 \\

\textbf{GEAR (Ours)}
& \textbf{0.537}
& \textbf{0.938} 
& \textbf{0.375}
& \textbf{0.453}
& \textbf{0.808} & \textbf{0.514} & \textbf{0.542} & \textbf{0.947} & \textbf{0.842} & 0.740 \\

\midrule
\multicolumn{11}{l}{\textit{Multi-Domain Generalization Evaluation (Qwen3-4B)}}\\
GRPO & 0.315 & 0.911 & 0.307 & 0.334 & 0.643 & 0.453 & 0.525 & 0.933 & 0.833 & 0.758 \\
\textbf{GEAR (Ours)}
& \textbf{0.343}
& \textbf{0.919}
& \textbf{0.325}
& \textbf{0.364}
& \textbf{0.724} & \textbf{0.533} & \textbf{0.623} & \textbf{0.952} & \textbf{0.849} &\textbf{ 0.776} \\

\bottomrule
\end{tabular*}}
\vspace{-10pt}
\end{table*}


\subsection{Ablation Study}
\label{ssec:ablation}

We conduct ablations to disentangle the contributions of GEAR's key components: (i) the granularity-adaptive mechanism for coherent credit assignment; and (ii) key hyperparameter settings.




\paragraph{Effectiveness of Dynamic Structured Segmentation.}
To analyze the contribution of our granularity-adaptive design in GEAR, we compare the full method against several segmentation variants in Table~\ref{tab:ablation}.
\textbf{w/o segment-level reweighting} removes segment-wise credit propagation and applies token-level KL-based reweighting independently to each token.
\textbf{w/o entropy termination} constructs segments solely based on KL triggers: each token with $\widetilde{\mathrm{rKL}}>0.1$ starts a new segment, which extends until the next KL trigger. All tokens within a segment share the same weight, determined by the first token's KL value.
\textbf{w/o KL-trigger} replaces KL-triggered segmentation with entropy-based boundaries: a segment terminates when token entropy exceeds $1.5\times$ the current entropy threshold, after which the next token initializes a new threshold. Segment-level weights are uniformly assigned according to the first token's KL value.
\textbf{with tool call segmentation} replaces our dynamic segmentation with tool-boundary segmentation, where each tool invocation starts a new segment, and all tokens within a segment share the weight determined by the first token's KL value.

All segmentation variants underperform GEAR, validating the effectiveness of our dynamic structured segmentation. In particular, removing segment-level reweighting (\textit{w/o segment-level reweighting}) shows that directly applying token-level KL modulation fails to propagate credit over coherent reasoning spans. Replacing our KL-initiated, entropy-terminated segmentation with single-signal partitioning (\textit{w/o entropy termination} or \textit{w/o KL-trigger}) further confirms the complementary roles of reverse KL and entropy: reverse KL identifies behavioral transitions, while entropy adaptively controls segment granularity. Segmenting trajectories solely by tool-calling boundaries (\textit{with tool call segmentation}) performs worst, suggesting that external structural boundaries are often misaligned with intrinsic reasoning dynamics. Overall, these results highlight the importance of behavior-aware segmentation and adaptive segment-level credit propagation for effective credit assignment.

\begin{wraptable}{rh}{0.5\textwidth}
\vspace{-10pt}
\centering
\small
\caption{Ablation study of GEAR. We evaluate the contribution of each component and the sensitivity to key hyperparameters. The default hyperparameter values are $\lambda_H=1.5$, $\alpha=0.2$, and $\lambda_{\mathrm{KL}}=0.1$.}
\label{tab:ablation}
\vspace{8pt}
\renewcommand{\arraystretch}{1.08}
\setlength{\tabcolsep}{4pt}
\begin{tabular}{lcc}
\toprule
Method Variant & Test & MATH \\
\midrule
GRPO baseline & 0.676 & 0.836 \\
\midrule
GEAR & \textbf{0.739} & \textbf{0.879} \\
\quad w/o segment-level reweighting & 0.683 & 0.844 \\
\quad w/o entropy termination & 0.691 & 0.845 \\
\quad w/o KL-trigger & 0.685 & 0.856 \\
\quad with tool call segmentation & 0.682 & 0.840 \\
\midrule
Entropy threshold $\lambda_H=1.3$ & 0.718 & 0.862 \\
Entropy threshold $\lambda_H=1.7$ & 0.693 & 0.851 \\
\midrule
Scale $\alpha=0.4, \beta=0.8$ & 0.733 & 0.872 \\
Scale $\alpha=0.6, \beta=0.7$ & 0.724 & 0.864 \\
KL threshold $\lambda_{\mathrm{KL}}=0.2$ & 0.693 & 0.854 \\
KL threshold $\lambda_{\mathrm{KL}}=0.3$ & 0.686 & 0.847 \\
\bottomrule
\end{tabular}
\vspace{-8pt}
\end{wraptable}

\textbf{Hyperparameter Analysis.}
We further examine the sensitivity of GEAR to the segmentation thresholds around the default settings.
\textit{For the entropy termination threshold}, smaller values ($\lambda_H=1.3$) produce finer-grained partitions by terminating segments earlier, while larger values ($\lambda_H=1.7$) encourage longer, more stable segments. Empirically, the default choice $\lambda_H=1.5$ provides the best balance between segmentation sensitivity and structural consistency, yielding segments that are neither overly fragmented nor excessively coarse. \textit{For the KL triggering threshold}, larger values ($\lambda_{\mathrm{KL}}=0.2$ or $0.3$) make segmentation overly conservative by requiring stronger deviation signals to start new segments. In contrast, the default setting $\lambda_{\mathrm{KL}}=0.1$ better captures localized behavioral shifts, enabling earlier transition detection and more precise credit assignment. Since occasional reverse-KL spikes can compress most normalized scores into a low range, a smaller threshold is preferable for maintaining sensitivity to informative deviations.

\textbf{Effectiveness of advantage modulation.}
We further study the modulation strength for reweighting trajectory-level advantage ($\alpha=0.4$ or $0.6$). The default coefficient ($\alpha=0.2$) yields the best trade-off between modulation expressiveness and optimization stability, validating our advantage correction design. By refining coarse trajectory-level supervision with structured token-level weighting, GEAR enables more informative credit propagation while preserving GRPO's stable optimization dynamics.

\subsection{Visualization Analysis of Segment Design}
\label{ssec:vis_segment}


Additional visualizations in the appendix
(Figures~\ref{fig:kl_tokens} and~\ref{fig:experiment_vis})  show that 
directly using token-level KL as a weighting signal mainly suppresses early non-reasoning tokens, which exhibit large distribution shifts but carry limited semantic relevance, leading to overly coarse weighting. When used for segmentation, KL produces sparse but high-amplitude spikes that dominate boundary detection and merge long reasoning spans into large segments. In contrast, entropy exhibits dense local fluctuations, resulting in fragmented and noisy segments that often break coherent reasoning structures. These results suggest that reverse KL captures salient deviation points, whereas entropy better reflects local uncertainty, highlighting their complementary roles in fine-grained credit assignment.

\section{Conclusion}
In this paper, we propose GEAR, a fine-grained credit assignment framework for reinforcement learning in LLM agents that transforms coarse trajectory-level supervision into adaptive segment- and token-level learning signals through self-distillation. By integrating self-distillation-based discrepancy estimation, dynamic trajectory segmentation, and outcome-aware advantage reweighting under the unified objective of fine-grained credit assignment, GEAR enables more precise and effective credit propagation over long-horizon trajectories without additional supervision. Extensive experiments on eight mathematical reasoning and agentic tool-use benchmarks demonstrate consistent improvements over strong GRPO-based baselines, highlighting the promise of coupling reference-guided signals with adaptive credit regions for scalable and generalizable agent reinforcement learning.

\clearpage
\bibliographystyle{unsrt}
\bibliography{reference}

\newpage
\appendix     
\section{Pseudocode}
\label{sec:Pseudocode}
Algorithm \ref{alg:GEAR} presents the proposed GEAR framework. Given a batch of prompts, the policy first generates trajectories and computes trajectory-level advantages. GEAR then performs divergence-aware segmentation based on reverse KL divergence and entropy signals, identifying segments dynamically. Within each trajectory, the advantage signal is redistributed into segment-level or token-level credits depending on the detected structure. Finally, an advantage-aware reweighting scheme is applied to modulate the learning signal, producing fine-grained, stable, and adaptive credit assignments for policy optimization. 

\begin{algorithm}[tb]
\caption{Granularity-AdaptivE Advantage Reweighting  (GEAR)}
\label{alg:GEAR}
\small
\begin{algorithmic}[1]
\Require Initial policy $\pi_{\theta}$, reference policy $\pi_{\mathrm{ref}}$, training batch $\mathcal D$, KL threshold $\lambda_{\mathrm{KL}}$, entropy threshold $\lambda_H$, scaling factors $\alpha,\beta$
\Ensure Updated policy parameters $\theta$

\For{each prompt $q_i \in \mathcal D$}
    \State Generate trajectory $\tau_k=\{(s_t,a_t)\}_{t=1}^{T_k}$ under $\pi_{\theta}$
    \State Compute trajectory-level advantage $A^{(k)}$
    
    \For{each step $t=1,\dots,T_k$}
        \State Compute reverse KL divergence:
        $\mathrm{rKL}_t=
        \log \pi_\theta(a_t\mid s_t)
        -
        \log \pi_\theta(a_t\mid s_t^\star)$
        \State Normalize $\mathrm{rKL}_t$ to obtain $\widetilde{\mathrm{rKL}}_t\in[0,1]$
        \State Compute token entropy $H_t$
    \EndFor
    
    \Statex \textbf{Divergence-aware segmentation}
    \State Initialize segment set $\mathcal S\gets\varnothing$, $t\gets 1$
    
    \While{$t\le T_k$}
        \If{$\widetilde{\mathrm{rKL}}_t>\lambda_{\mathrm{KL}}$}
            \State Set segment onset $t_{\mathrm{start}}\gets t$
            \State Find first $t_{\mathrm{end}}$ satisfying $H_{t_{\mathrm{end}}}>
            \lambda_H H_{t_{\mathrm{start}}}$

            \State Add segment $[t_{\mathrm{start}},t_{\mathrm{end}}]$ to $\mathcal S$
            \State $t\gets t_{\mathrm{end}}+1$
        \Else
            \State $t\gets t+1$
        \EndIf
    \EndWhile
    
    \Statex \textbf{Piecewise credit assignment}
    \For{each step $t=1,\dots,T_k$}
        \If{$t\in S_n,\; S_n\in\mathcal S$}
            \State Assign segment-level credit: 
            $w_t^{\mathrm{KL}}
            =
            \widetilde{\mathrm{rKL}}_{t_n}$
        \Else
            \State Assign token-level credit
            $w_t^{\mathrm{KL}}
            =
            \widetilde{\mathrm{rKL}}_t$
        \EndIf
        
        \State Compute sign-aware weight
        $w_t=
        0.5+
        (0.5-w_t^{\mathrm{KL}})
        \cdot
        \mathrm{sign}(A^{(k)})$
        
        \State Compute final modulation weight
        $W_t=\alpha w_t+\beta$
        \State Assign reweighted advantage
        $\hat A_t^{(k)}=W_tA^{(k)}$
    \EndFor
\EndFor

\State Update policy $\theta$ with
\begin{equation}
\label{eq:final_grpo_objective}
\mathbb{E}\Bigg[
\frac{1}{K}
\sum_{k=1}^{K}
\frac{1}{|\tau^{(k)}|}
\sum_{t=1}^{|\tau^{(k)}|}
\min\!\Big(
r_t^{(k)}(\theta) \hat{A}_t^{(k)},
\mathrm{clip}\!\big(r_t^{(k)}(\theta),1-\epsilon,1+\epsilon\big) \hat{A}_t^{(k)}
\Big)
\Bigg]
- \beta D_{\mathrm{KL}}(\pi_\theta \| \pi_{\mathrm{ref}}),
\end{equation}

\end{algorithmic}
\end{algorithm}

\section{Implementation Details}
\label{sec:Implementation Details}
\subsection{Training Setup}

We implement our method based on VolcEngine Reinforcement Learning (VeRL) \citep{sheng2025hybridflow} and train it with a learning rate of $1\times10^{-6}$, a batch size of $128$, and $8$ rollouts per prompt for $2$ epochs on \textsc{NVIDIA H100} GPUs. Unless otherwise specified, all other hyperparameters follow the default GRPO settings.

\subsection{Instruction Prompt for Training data}

\paragraph{Mathematical Reasoning.} All math tasks use the following instruction:
\begin{Verbatim}[
breaklines=true,
breakanywhere=true,
breaksymbolleft={},
breaksymbolright={},
fontsize=\small
]
You are a helpful assistant that can solve the given question step by step with the help of the python interpreter tool. Given a question, you need to first think about the reasoning process in the mind and then provide the answer. During thinking, you can invoke python interpreter tool to calculate the math problem for fact information about specific topics if needed. The reasoning process and answer are enclosed within <think> <\/think> and <answer> <\/answer> tags respectively. For example, <think> This is the reasoning process. <\/think> <think> This is the reasoning process. <\/think> <python> python code here <\/python> <result> python interpreter result here <\/result> <think> This is the reasoning process. <\/think> <answer> The final answer is \\[ \boxed{answer here} \\] <\/answer>. In the last part of the answer, the final exact answer is enclosed within \boxed{} with latex format.
\end{Verbatim}

\paragraph{Agentic Tool-use.}

An example is shown below:
\begin{Verbatim}[
breaklines=true,
breakanywhere=true,
breaksymbolleft={},
breaksymbolright={},
fontsize=\small
]
You are a helpful multi-turn dialogue assistant capable of leveraging tool calls to solve user tasks and provide structured chat responses.\n\n**Available Tools**\nIn your response, you can use the following tools:\n1. Name: checkMembership\nDescription: Check if a person is a member of the library and has access to the library services\nParameters: {"user_id": {"description": "The unique identifier of the library user (e.g., library card number, username)", "type": "string", "default": ""}, "pin": {"description": "The personal identification number of the library user", "type": "string", "default": ""}}\n2. Name: logAccessEvent\nDescription: Log an access event within the library for auditing and security purposes\nParameters: {"user_id": {"description": "The unique identifier of the library user (e.g., library card number, username)", "type": "string", "default": ""}, "event_type": {"description": "The type of access event (e.g., entry, exit, resource access)", "type": "string", "default": ""}}\n3. Name: authorizeEntry\nDescription: Authorize entry of a person into the library premises\nParameters: {"user_id": {"description": "The unique identifier of the library user (e.g., library card number, username)", "type": "string", "default": ""}, "pin": {"description": "The personal identification number of the library user", "type": "string", "default": ""}}\n4. Name: getLibraryAccessControl\nDescription: Check the access control settings in a library\nParameters: {"library_name": {"description": "The name of the library you want to check the access control", "type": "string", "default": ""}, "user_id": {"description": "The ID of the user requesting access control information", "type": "string", "default": ""}, "time_of_day": {"description": "Specify a time of day for access control (e.g., morning, afternoon, evening)", "type": "string", "default": ""}}\n\n**Steps for Each Turn**\n1. **Think:** Recall relevant context and analyze the current user goal.\n2. **Decide on Tool Usage:** If a tool is needed, specify the tool and its parameters.\n3. **Respond Appropriately:** If a response is needed, generate one while maintaining consistency across user queries.\n\n**Output Format**\n```plaintext\n<think> Your thoughts and reasoning <\/think>\n<tool_call>\n{"name": "Tool name", "parameters": {"Parameter name": "Parameter content"}}\n<\/tool_call>\n<think> Your thoughts and reasoning for the next tool call <\/think>\n<tool_call>\n{"name": "Tool name", "parameters": {"Parameter name": "Parameter content"}}\n<\/tool_call>\n...\n<response> AI's final response <\/response>\n```\n\n**Important Notes**\n1. You must always include the <think> field to outline your reasoning. Provide at least one of <tool_call> or <response>. Decide whether to use <tool_call> (possibly multiple times), <response>, or both.\n2. Do NOT invoke multiple tool calls simultaneously. Use exactly one tool call at a time. Before every tool call, provide a dedicated <think> step for that single call. Each tool call should be a JSON object with a "name" field and a "parameters" field. If no parameters are needed, keep the "parameters" field as an empty dictionary.\n3. Refer to the previous dialogue records in the history, including the user's queries, previous <tool_call>, <response>, and any tool feedback noted as <obs> (if exists).
\end{Verbatim}

\section{Additional Results and Analysis}
\label{sec:more_experiment_results}

\subsection{Analysis of Token-Level Reverse KL Divergence}

\begin{figure}
\vspace{-8pt}
\centering
\includegraphics[width=0.8\linewidth]{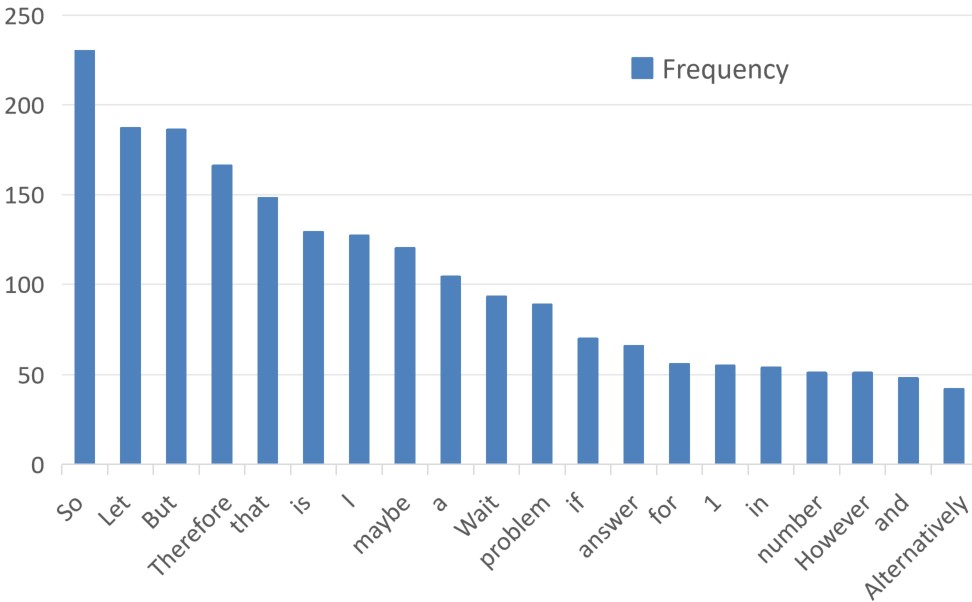}
\caption{Frequency of top-20 tokens with normalized reverse-KL $>0.1$ in 100 sampled math trajectories.}
\label{fig:kl_tokens}
\end{figure}

To better understand the behavior of reverse KL signals, we randomly sample 100 trajectories and compute the reverse KL divergence between a ground-truth-conditioned teacher policy and the corresponding student policy over student-generated responses. We then normalize token-level reverse KL values and analyze the top-20 most frequent high-KL tokens (normalized reverse-KL $>0.1$) across trajectories 
(Figure~\ref{fig:kl_tokens}).

We observe that reverse KL is highly concentrated on a small number of tokens, often peaking at the first few tokens of a segment. These high-amplitude spikes tend to dominate boundary detection, which can merge long reasoning spans into overly coarse segments. As a result, if credit assignment is performed solely based on these high-KL tokens, the learning signal becomes overly sparse and biased, while neglecting the internal structure of subsequent reasoning steps.

\subsection{Analysis of Segment Triggering and Termination}

\begin{figure}
\vspace{-4pt}
\centering
\includegraphics[width=0.85\linewidth]{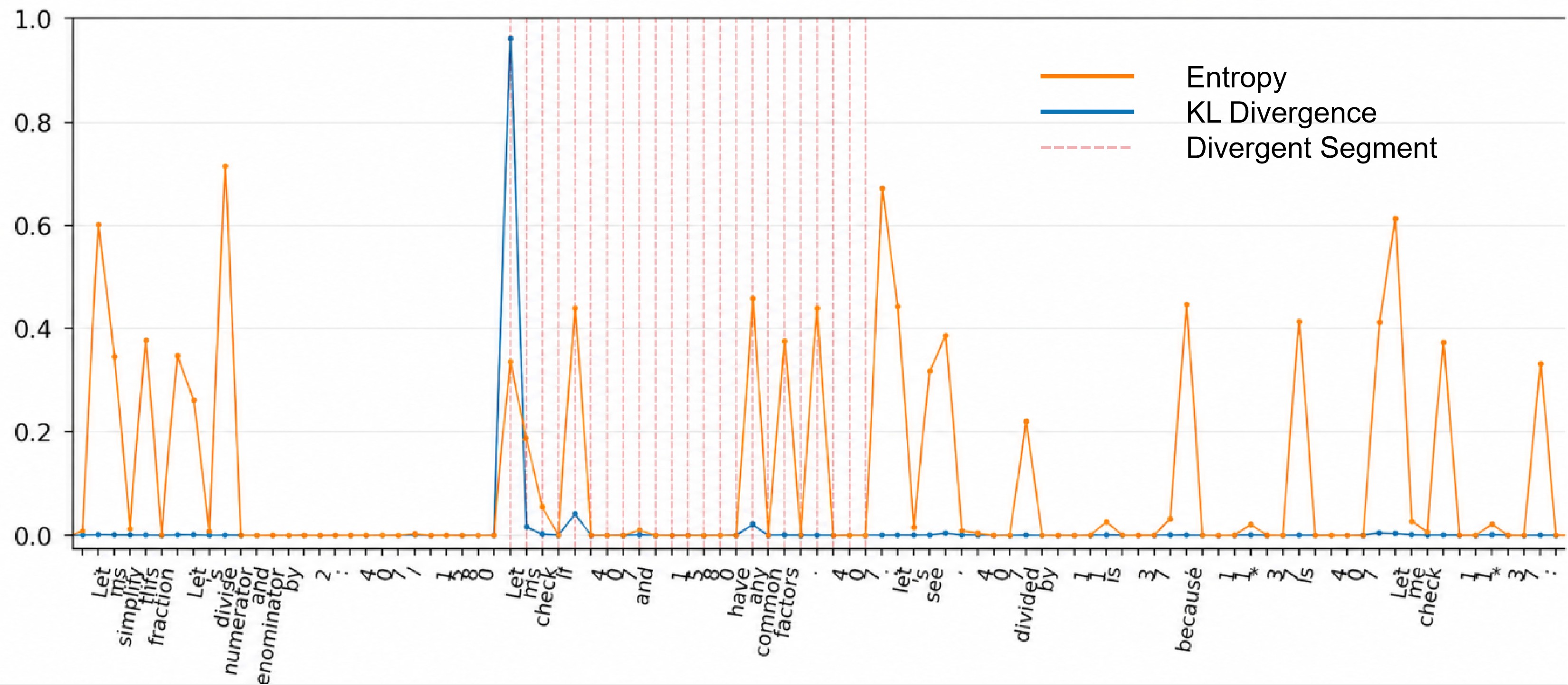}
\caption{Token-level visualization results of normalized KL divergence and normalized entropy.}
\label{fig:experiment_vis}
\end{figure}

Figure~\ref{fig:experiment_vis} reveals distinct limitations of reverse KL and entropy when used independently for credit assignment and segmentation. We observe that reverse KL is highly sparse and typically concentrates on the beginning tokens of a divergent reasoning span, producing sharp high-amplitude spikes. While such spikes effectively capture the onset of behavioral deviation, directly using token-level KL as a weighting signal mainly suppresses early non-reasoning or transition tokens that exhibit large distribution shifts but carry limited semantic relevance, resulting in overly coarse token weighting. When used for segmentation, these sparse spikes dominate boundary detection, often triggering segment boundaries too early and merging subsequent coherent reasoning steps into large segments, which may lead to inaccurate segmentation and insufficient credit differentiation within the reasoning span. 

In contrast, entropy exhibits dense local fluctuations throughout generation. Although entropy better reflects local uncertainty and semantic transitions, its high sensitivity to token-level variation often produces fragmented and noisy segments that break coherent reasoning structures into many short spans, introducing unstable and overly localized supervision signals. 

Taken together, these results suggest that reverse KL provides a sharp but temporally sparse signal of deviation onset, whereas entropy offers a dense but noisy estimate of local behavioral transitions. Their complementary characteristics motivate our adaptive segmentation design, where reverse KL is used to identify meaningful deviation triggers and entropy is used to determine segment extent, enabling more reliable fine-grained credit assignment.

\subsection{Entropy Termination}

We further conduct an experiment replacing token-level entropy termination with an entropy-window strategy. Specifically, instead of using the entropy of the current token to determine segment termination, we compute the average entropy over a sliding window of 8 tokens while applying the same threshold (\(\lambda_H = 1.5\)). As shown in Table~\ref{tab:entropy_window}, this variant underperforms the original token-level design. We hypothesize that reasoning shifts and behavioral deviations in long-horizon trajectories are often triggered by a small number of critical tokens. Averaging entropy across multiple tokens smooths these sharp local fluctuations and delays boundary detection. Consequently, the resulting segment boundaries become less aligned with the underlying reasoning dynamics, leading to coarser segmentation and less precise credit assignment.

\begin{table}
\vspace{-5pt}
\centering
\small
\caption{Ablation study of GEAR. The default hyperparameter values are $\lambda_H=1.5$, $\alpha=0.2$, and $\lambda_{\mathrm{KL}}=0.1$.}
\label{tab:entropy_window}
\vspace{8pt}
\renewcommand{\arraystretch}{1.08}
\setlength{\tabcolsep}{4pt}
\begin{tabular}{lcc}
\toprule
Method Variant & Test & MATH \\
\midrule
GRPO baseline & 0.676 & 0.836 \\
\midrule
GEAR & \textbf{0.739} & \textbf{0.879} \\
\quad with entropy window termination & 0.687 & 0.848 \\
\bottomrule
\end{tabular}

\end{table}

\subsection{Training Curve}

\begin{figure}
\vspace{-5pt}
\centering
\includegraphics[width=\linewidth]{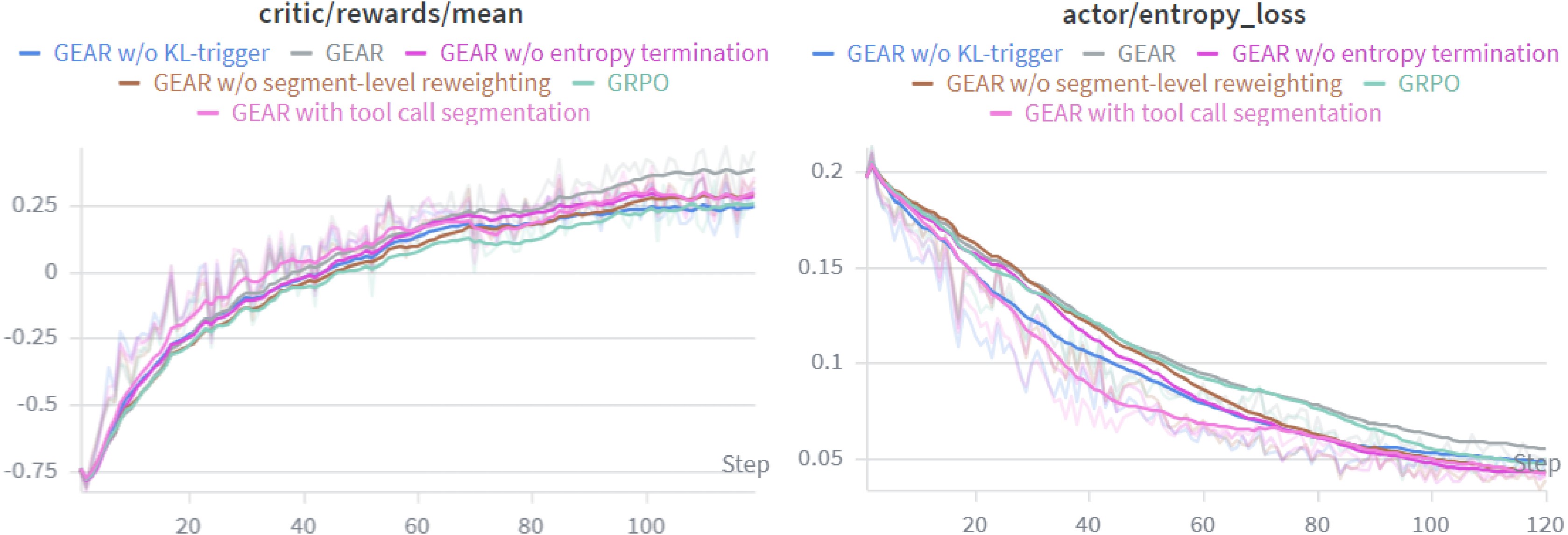}
\caption{Training curves of GRPO, GEAR and its variants. The left panel shows the mean training reward, while the right panel illustrates the policy entropy. GEAR achieves the highest training reward while maintaining higher policy entropy compared to the GRPO baseline and other ablated versions.}
\label{fig:training_curve}
\end{figure}

From the training curves (Figure~\ref{fig:training_curve}), GEAR achieves both higher reward and stronger exploration during RL optimization. GEAR converges to the highest average reward, while removing any key component leads to noticeable degradation, validating the necessity of the full design. Meanwhile, GEAR maintains higher policy entropy throughout training, indicating stronger behavioral diversity and reduced premature policy collapse. We attribute these gains to GEAR's fine-grained credit assignment mechanism, which adaptively allocates trajectory-level advantage to behaviorally meaningful tokens and segments rather than uniformly propagating rewards across the entire trajectory. This more precise and outcome-consistent credit propagation suppresses noisy credit assignment, and preserves useful behavioral variation, enabling more stable policy evolution and avoiding early convergence to suboptimal behaviors.

\subsection{Computational Cost}
\label{ssec:Computational}
For mathematical reasoning tasks, training our method for one epoch on the 4B model requires approximately 26 hours using 8 H100 GPUs, which is comparable to the training time of the GRPO baseline. For agentic tool-use tasks, our method requires approximately 10 hours per epoch under the same hardware setting, also comparable to GRPO.

\section{Limitation}
\label{sec:Limitation}
GEAR relies on access to ground-truth answers to construct the teacher policy conditioned on the ground truth. This design choice is well-aligned with supervised settings such as mathematical reasoning and tool-use benchmarks with annotated reference trajectories, where reliable ground-truth signals are available. In more open-ended scenarios, where verifiable ground truth may be ambiguous or multiple valid solutions exist, similar supervision signals can potentially be obtained through alternative evaluative mechanisms. In such cases, the reverse KL signal may be defined using proxy judgments, such as learned verifiers or preference-based evaluation criteria, rather than explicit ground-truth annotations. Future work could explore these directions to extend GEAR to broader settings without requiring direct access to ground-truth-conditioned teacher policies.

\end{document}